\documentclass[runningheads]{llncs}
\usepackage[T1]{fontenc}
\usepackage[american]{babel}
\usepackage{mathtools}
\usepackage{amssymb}
\usepackage{graphicx}
\usepackage{booktabs}
\usepackage{subcaption}
\usepackage{xcolor}
\usepackage{float}
\newcommand{\localRel}{e_l} 
\newcommand{\localRelNC}{e_{l,\nc}}
\newcommand{\globalRelUNC}{e_{\unc}}
\newcommand{\globalRelCER}{e_{\cer}}

\newcommand{\model}{\mathbf{M}}
\newcommand{\X}{\mathbf{X}}
\newcommand{\x}{\mathbf{x}}
\newcommand{\W}{\mathbf{W}}
\newcommand{\V}{\mathbf{V}}
\newcommand{\A}{\mathbf{A}}
\newcommand{\M}{\mathbf{M}}
\newcommand{\data}{\mathcal{D}}
\newcommand{\thet}{\boldsymbol{\theta}}
\newcommand{\ucurr}{u}

\newcommand{\utot}{u_{t}}
\newcommand{\uep}{u_{e}}
\newcommand{\ual}{u_{a}}

\newcommand{\utoth}{\hat{u}_{t}}
\newcommand{\ueph}{\hat{u}_{e}}
\newcommand{\ualh}{\hat{u}_{a}}

\newcommand{\Hentr}{\mathcal{H}}
\newcommand{\cer}{\text{CER}}
\newcommand{\unc}{\text{UNC}}
\newcommand{\nc}{nc}

\begin{document}
\title{\textit{Concept}ualizing Uncertainty: A \textit{Concept}-based Approach to Explaining Uncertainty}
\titlerunning{\textit{Concept}ualizing Uncertainty}
%

\author{Isaac Roberts\and
Alexander Schulz\and
Sarah Schr\"{o}der\and
Fabian Hinder\and
Barbara Hammer}

\authorrunning{I. Roberts et al.}
%

\institute{Machine Learning Group, Bielefeld University, D-33619 Bielefeld, Germany
\email{\{iroberts,aschulz,saschroeder,fhinder,bhammer\}@techfak.uni-bielefeld.de}}

\authorrunning{I. Roberts et al.}
%
%
\maketitle              
\begin{abstract}
Uncertainty in machine learning refers to the degree of confidence or lack thereof in a model’s predictions. While uncertainty quantification methods exist, explanations of uncertainty, especially in high-dimensional settings, remain an open challenge. 
Existing work focuses on feature attribution approaches which are restricted to local explanations.
Understanding uncertainty, its origins, and characteristics on a global scale is crucial for enhancing interpretability and trust in a model's predictions. In this work, we propose to explain the uncertainty in high-dimensional data classification settings by means of concept activation vectors, which give rise to local and global explanations of uncertainty. We demonstrate the utility of the generated explanations by leveraging them to refine and improve our model. \footnote{Code is freely available here: \url{https://github.com/robertsi20/Conceptualizing-Uncertainty}.} 

\keywords{Explainable Uncertainty\and Concept-based Explanations\and XAI.}
\end{abstract}
\section{Introduction}

\begin{figure}[t]
\includegraphics[width=\textwidth]{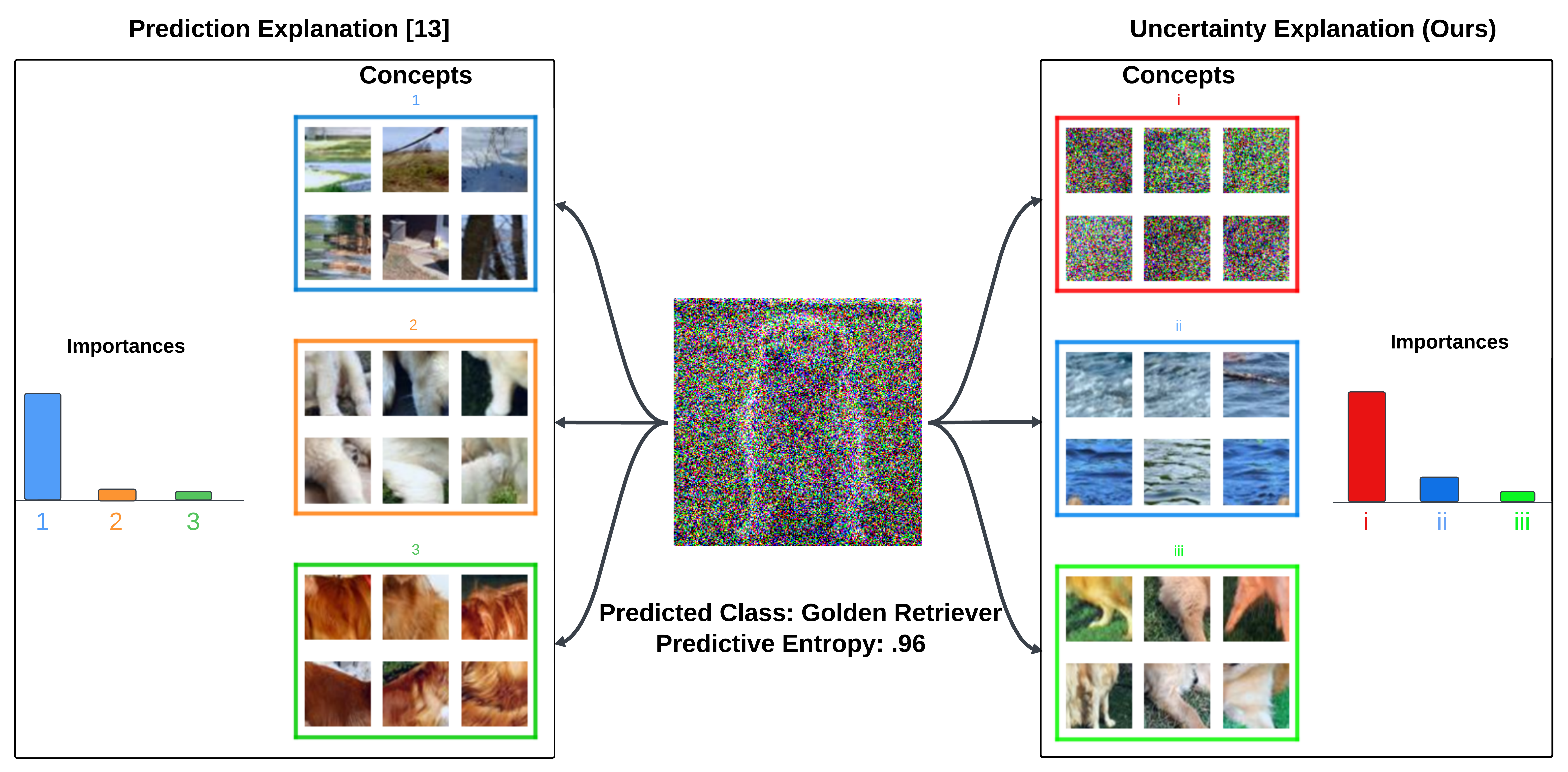}
\caption{
We display concept-based explanations of a model \emph{prediction} from literature \cite{fel2023craft} (\textbf{left}) and our proposed concept-based explanations of \emph{predictive uncertainty} (\textbf{right}) for the same input. Each explanation contains three most important concepts, each visualized by the 6 most activating patches from the training set. 
The prediction explanation suggests the neural net arrived at its decision primarily due to the detected concept {\color{blue} 1} of the background, while the uncertainty explanation offers that the model is uncertain about its prediction largely because of the detected concept {\color{red} i} of the noise.} \label{fig:example}
\end{figure}

While advances in deep learning in recent years have led to impressive performance in many domains, such models are not always reliable and pose risks in real-world applications, especially when exposed to dynamic environments. As such, numerous methods have been developed in the field of explainable artificial intelligence (xAI) \cite{burkart2021XAIsurvey} to provide insights into model behavior and facilitate actionable modifications. However, most methods focus on explaining model \emph{predictions}, 
which does not explicitly address predictive \emph{uncertainty} (see Figure \ref{fig:example}). Understanding sources of uncertainty is crucial for detecting potential model weaknesses and data flaws and, additionally, provides means of meaningful downstream actions \cite{H_llermeier_2021}, aimed at increasing trust and reliability.

Understanding uncertainty and its sources requires 3 main steps: 1. localizing it, 2. assigning a degree of uncertainty, and 3. finding its origin. As such, Uncertainty Quantification (UQ) methods emerged as a tool and have proven useful in various applications, including active learning \cite{kirsch2019batchbald}, classification with rejects \cite{hendrickx2024machine}, adversarial example detection \cite{DBLP:conf/uai/SmithG18}, reinforcement learning \cite{osband2016deep}, and separating sources of uncertainty \cite{H_llermeier_2021}. Therefore, a significant body of work aims to improve the quantification of predictive uncertainty using techniques such as Bayesian deep learning (BDL) and approximations thereof \cite{gawlikowski2023surveyUncert,depeweg2018decomposition,gal2016dropout}. However, existing methods of UQ assign values of certainty without providing human-interpretable insights about what may be causing it \cite{gawlikowski2023surveyUncert}.

For this reason, some attempts have been made to explain uncertainty with state-of-the-art xAI techniques, including feature attributions \cite{watson2024ExplWShap,wang2023ExplUABackp} and counterfactuals \cite{antoran2020ExplCounterf}. Although these approaches provide valuable information for individual data points, they are restricted to local explanations and do not provide high-level information on a global dataset scale, such as in image datasets. Additionally, feature attribution methods are limited, because they point only to \emph{which} part of the input the model considers as important, but do not explain \emph{why} \cite{fel2023craft}. In this regard, methods aiming to automatically extract concept activation vectors (CAVs) \cite{fel2024holistic,achtibat2023attribution} and attribute them back to the input provide both information on what the model seems to perceive in a part of the input and global dataset information, by inspecting the learned concepts.

In this work, (i) we propose a novel pipeline to enable concept-based explanations for predictive uncertainty, providing both local and global explanations for sources of uncertainty in a human-interpretable way, and (ii) we demonstrate the potential to perform actionable interventions based on the learned concepts in a series of downstream tasks, including the automatic detection of different types of uncertainty in new environments, interpretable uncertainty-based rejections and detecting gender bias in language models, thereby showing their usefulness.


This paper is organized as follows: Section \ref{sec:background} provides background information needed to understand our proposed pipeline as well as related work. In Section \ref{sec:pipeline}, we provide the details of our pipeline. Section 4 presents the experiments where we evaluate the pipeline's effectiveness. We finish our work with a discussion on limitations in Section \ref{sec:lim} and a conclusion.

\section{Background} \label{sec:background}

In this section, we list the relevant fundamentals, including our problem formulation, background on uncertainty quantification and concept activation vectors, as well as related work on explaining uncertainty.

\subsection{Setup and Problem Formulation} \label{subsec:problem}

Given a trained classification model $\model$, e.g.,\ a deep convolutional network, and a dataset of $n$ data points $\X = \{\x_1, ..., \x_n \}$, usually not seen during training. In contrast to a vast amount of literature focusing on explaining predictions of such models \cite{burkart2021XAIsurvey}, we aim to explain their predictive uncertainty, thereby aiming to understand its sources.
Our work differs from recent research on that topic, e.g.,\ \cite{burkart2021XAIsurvey}, in that we aim for human-interpretable concept-based explanations. 

For the following UQ, we proceed with a Bayesian formalization due to its precedence \cite{gawlikowski2023surveyUncert,H_llermeier_2021} in the literature. As such, we also require the training data set $\data$ of $\model$, which, however, our method does not use. \footnote{Other notable frameworks include Conformal Prediction \cite{10.5555/1390681.1390693} and Frequentist approaches \cite{gawlikowski2023surveyUncert} and could also be used with our pipeline. } 


\subsection{Quantifying Uncertainty}\label{subsec:est_unc}

A popular way to define predictive uncertainty is over the predictive distribution $p(y|\x, \data)$ \cite{gawlikowski2023surveyUncert,depeweg2018decomposition},  which in our case is over possible labels $y$.

 In our Bayesian setting, where the parameters $\thet$ of our classification model $\model$ are random variables, $p(y|\x, \data) = \mathbb{E}_{p(\thet|\data)}[p(y|\x,\thet)]$ requires computing the expectation $\mathbb{E}$ over the posterior $p(\thet|\data)$, which is usually intractable.
Accordingly, various methods for its approximation have been introduced \cite{gawlikowski2023surveyUncert}, such as Variational Inference based approaches like Monte Carlo (MC) dropout \cite{gal2016dropout}, sampling based methods \cite{welling2011bayesian}, Laplace Approximations \cite{NEURIPS2021_a7c95857} and ensembles \cite{lakshminarayanan2017QuantifyUncert}.


Different measures for predictive uncertainty are defined in the literature \cite{gawlikowski2023surveyUncert,depeweg2018decomposition}. We summarize and use the measures in the following due to their prominence in the literature \cite{watson2024ExplWShap,wang2023ExplUABackp}, but our explanation approach can be applied with any measure. We can also compute metrics that quantify uncertainty sources into their aleatoric and epistemic components. Since this is a classical way of explaining sources of uncertainty, we include them for comparison to our method. \\ 

\emph{Total Uncertainty} based on Shannon Entropy:
\begin{equation}
    \utot(\x) \coloneq \Hentr [ p(y | \x, \data ) ] 
    = - \sum_{y \in Y} p(y | \x, \data) \log_2 p(y | \x, \data)
    \label{eq:totalU}
\end{equation}
\emph{Aleatoric} and \emph{Epistemic Uncertainty} based on the decomposition of $\utot$:
\begin{equation}
\ual(\x) \coloneq \mathbb{E}_{p(\thet|\data)}\big[\Hentr [p(y|\x,\thet)]\big], ~~~~~ \uep(\x) \coloneq \utot(\x) - \ual(\x)
\label{eq:aleaU}
\end{equation}


To approximate the above measures, we utilize MC dropout to collect a set of predictions $\{ p(y|\x,\thet_i)\}_{i=1}^N$ and approximate the posterior predictive
$p(y|\x, \data) = \mathbb{E}_{p(\thet|\data)}[p(y|\x,\thet)] \approx \frac{1}{N} \sum_i^N p(y|\x,\thet_i).$
We refer to $\utoth, \ualh, \ueph$ when utilizing this approximation in $\utot, \ual, \uep$, respectively.

\subsection{Concept Activation Vectors}

Concept Activation Vectors (CAVs) aim for human interpretability with respect to understanding black-box model predictions \cite{kim2018interpretabilityfeatureattributionquantitative,fel2024holistic,NEURIPS2019_77d2afcb}. These can be categorized into two classes, concept bottleneck models which enforce the use of concepts during training and post-hoc methods that are applied after training and provide additional information as compared to saliency maps \cite{fel2023craft}. In the present work, we focus on such post-hoc approaches, since concept bottleneck models usually require concept labels (apart of some notable exceptions \cite{oikarinen2023labelfree}) and we are interested in concepts that explain uncertainty.
Recent approaches based on Nonnegative Matrix Factorization (NMF) \cite{fel2023craft,cox_llm,jourdan-etal-2023-cockatiel} have been shown to demonstrate superior qualitative and quantitative properties of the resulting concepts \cite{fel2024holistic}. Here, (parts of) the input are typically embedded 
into a non-negative activation space of a pre-trained model and NMF decomposes the embedded data matrix $\A$ into a product of non-negative matrices $\W$ and $\V$,  solved by reconstructing $\A$, i.e., $(\W,\V) = \arg \min_{\W\geq 0, \V\geq 0} \| \A - \W\V^\top \|^2_F$.  The decomposition yields: $\V$ the dictionary of concepts (or concept bank) and $\W$ a reduced representation of $\A$ according to the basis $\V$. To attribute importance, the authors of \cite{fel2023craft} make use of a sensitivity analysis technique known as total Sobol Indices, which captures the effects of a concept along with its interactions on the model's output by considering the variance fluctuations by perturbing $\W$. The contribution of concept $i$ is then defined by: $ S_i^T = \mathbb{E}_{\M_{\sim i}} \left( \mathbb{V}_{\M_i} (h((\W \circ \M)\V^\top)|\M_{\sim i}) \right) \backslash \mathbb{V} \left( h((\W \circ \M)\V^\top) \right) $, where $h$ is the model function mapping embeddings $\A$ to the model's output, $\M$ are uniformly and i.i.d.\ stochastic masks in $[0,1]^r$ with $r$ concepts, $\circ$ the Hadamard product, $\sim$ the complementary function \cite{fel2023craft}.

\subsection{Related Work on Explaining Uncertainty}
\label{subsec:relWork}

While a plethora of xAI methods exists for explaining the prediction of classification models, including several local and global approaches \cite{burkart2021XAIsurvey}, methods for explaining the source of uncertainty have only developed recently. Mostly, these have focused on local \emph{feature attribution} explanations, including explaining uncertainty with shapley values \cite{watson2024ExplWShap}, with gradient-based methods
\cite{wang2023ExplUABackp}, with counterfactuals \cite{antoran2020ExplCounterf} and by taking second-order effects into account \cite{bley2025ExplSecondOrder}. 
In contrast, we aim for explanations beyond feature attribution that also provide global explanations of uncertainty, which enable an overview of uncertain prediction characteristics. 
Also, concerning image datasets, rather small ones consisting of MNIST and CIFAR are used in \cite{watson2024ExplWShap,wang2023ExplUABackp,antoran2020ExplCounterf}. Only \cite{bley2025ExplSecondOrder} is applied to CelebA, containing larger images.
We apply our approach to images of ImageNet.


A few methods have been proposed for explaining uncertainty on a global scale. These include \cite{unsupdeepview}, which utilizes dimensionality reduction and an adaptation of \cite{deepview} to visualize uncertainty patterns and \cite{conformasight}, using conformal prediction.

Another, less related line of work, aims to provide uncertainty estimates of explanations \cite{salvi2025explainability}, e.g., by utilizing uncertainty sets \cite{marx2023but}. We, in contrast, aim to explain the uncertainty of the classifier, not the uncertainty of an explanation.

\section{Proposed Pipeline} \label{sec:pipeline}

\begin{figure}[t]
\includegraphics[width=\textwidth]{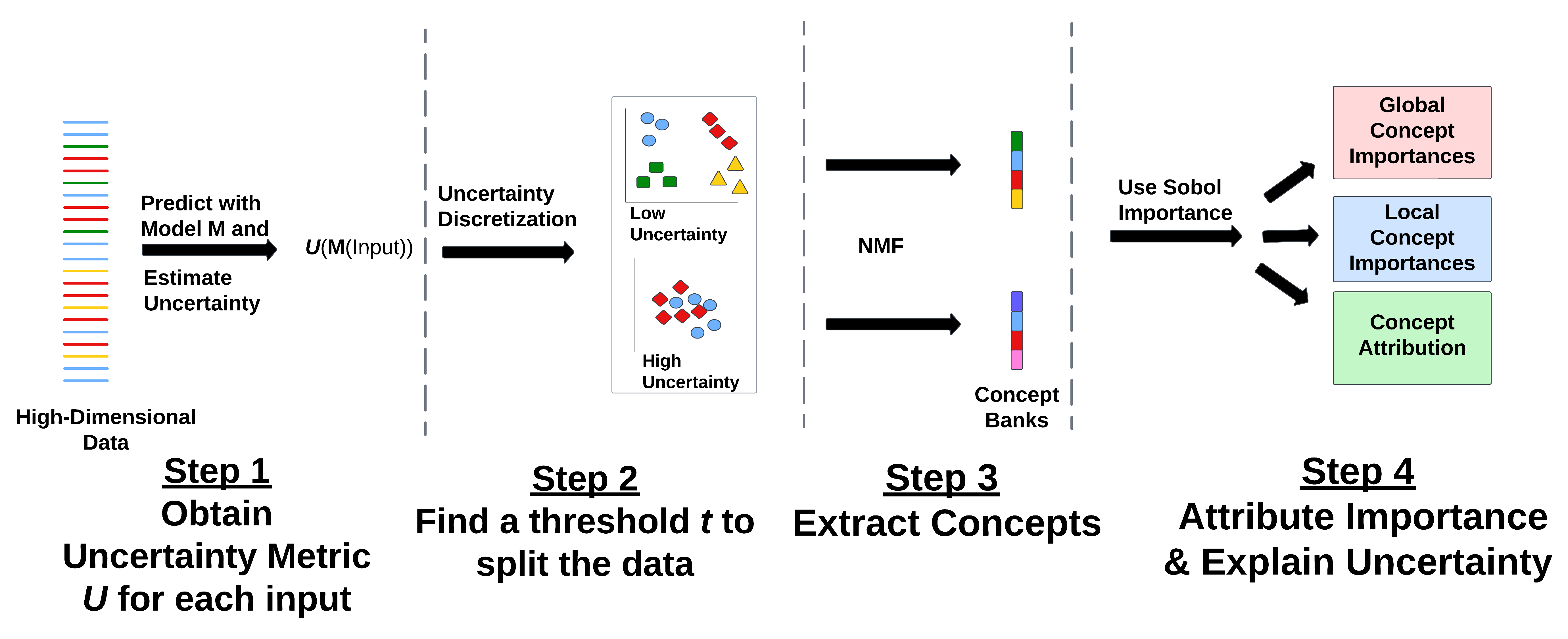}
\caption{Our proposed pipeline for uncertainty explanation using CAVs. } \label{fig:methodology}
\end{figure}



We propose to explain predictive uncertainty by means of CAVs computed on a local level -- for each data point -- or aggregated to obtain a global explanation. In Figure \ref{fig:methodology}, we illustrate our proposed pipeline, which aims to characterize and explain uncertainty using extracted concepts
from high-dimensional data $\X$, by grouping the predictions into certain and uncertain ones and extracting concepts from these. More specifically, given such data $\X$, a classification model $\model$, and an uncertainty measure $\ucurr$, e.g.,\ $\ucurr\coloneq\utoth$, we arrange our pipeline into 4 steps:

\noindent \textbf{Step 1} -
We use the model $\model$ to classify the inputs and compute $\ucurr(\x_i)$ for each data point, by leveraging an approximation technique such as MC Dropout.

\noindent \textbf{Step 2} - In order to group $\ucurr(\x_i)$ into uncertain (UNC) and certain (CER) samples, we specify a probabilistic classification task $f : \ucurr(\x) \mapsto [0,1]$, such that $f(\ucurr(\x)) < 0.5$ corresponds to CER samples.
We therefore expect that if applied to all data points, $\{\ucurr(\x_i)\}_{i=1}^n$ can be described by a mixture model with two components. For simplicity, we assume a Gaussian Mixture Model (GMM) with two components, which we train on $\{\ucurr(\x_i)\}_{i=1}^n$. By our assumption, we expect the component with the larger mean to correspond to the UNC samples. We thus obtain the classification model $f$ by considering the conditional probability, which takes on a sigmoid shape. 

\noindent \textbf{Step 3} - To generate the concepts, we embed the data using a foundation model $g$ into an activation space with the condition that for each $\x_i$, $g(\x_i) \geq 0 $ (e.g.,\ after a ReLU layer), and then we train one NMF on patches from $\{g(\x_i)| \x_i\in \unc \}$ and another NMF on patches from $\{g(\x_i)| \x_i\in \cer \}$, producing two concept banks, $\V_{\unc}$ and $\V_{\cer}$.
Thus, we can represent each $g(\x_i)$ as a linear combination of the concepts in $\V_\text{\unc}$ or $\V_\text{\cer}$, with scaling factors $\W_i$.

\noindent \textbf{Step 4a} - To estimate the importance of the concepts in $\V_\text{\unc}$ and $\V_\text{\cer}$, we utilize the Sobol Indices \cite{fel2021lookvarianceefficientblackbox,fel2023craft,jourdan-etal-2023-cockatiel}, using $f$ as the function of interest.


\noindent \textbf{Step 4b} -  Repeating Step 4a for every data point, we obtain a \emph{local} importance score $\localRel(g(\x_i)) \in \mathbb{R}^{d}$ with $d$ concepts. Additionally, we supplement the local importances with an attribution map \cite{fel2023craft,jourdan-etal-2023-cockatiel} indicating where the important concepts are detected in the input. We further augment the local explanations $\localRel$ with consistent global explanations. The \emph{global} importances can be computed as in literature by averaging over $\localRel(g(\x_i))$ for points predicted in $\unc$ and $\cer$, respectively, producing $\globalRelUNC$ and $\globalRelCER$. 





 \section{Experiments} \label{sec:experiments}

Since uncertainty arises from multiple sources, establishing a ground truth for experimentation can be challenging. Therefore, we aim to demonstrate the validity and usefulness of our uncertainty explanations by illustrating how our concepts capture different sources of uncertainty, aiding human decision-makers in constructing effective re-training sets. Additionally, we integrate them into a classification with reject options setting. 
Finally, we show that our concepts can reveal 
potential biases concerning sensitive groups in a downstream task.

\begin{figure}[t]
\includegraphics[width=\textwidth]{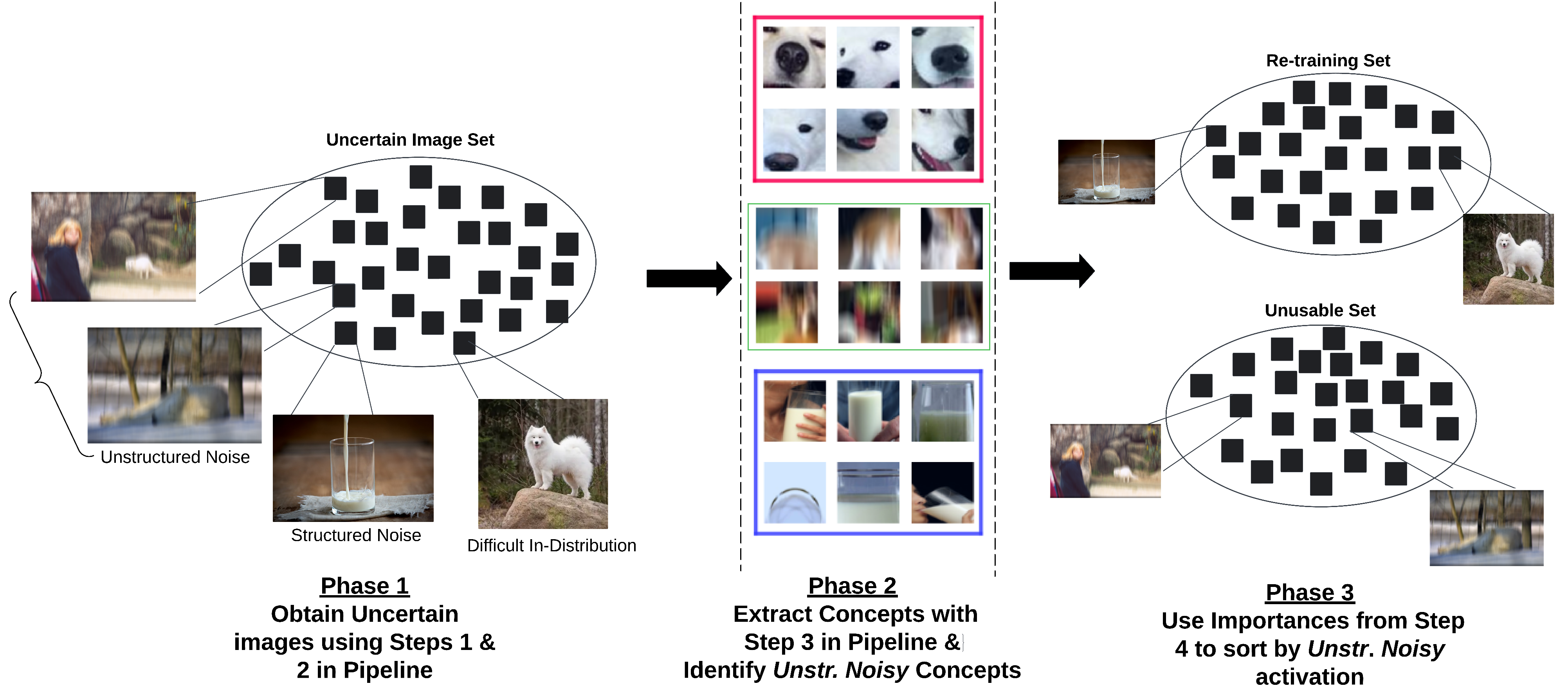}
\caption{Setup of Experiment \ref{subsec:distingUnc}: Points with high uncertainty (Phase 1) are used for concept extraction (Phase2). Concepts are inspected manually for unstructured noise (concept 2 is identified here). Uncertain points are grouped according to the activation of the noise concept automatically (Phase 3).} \label{fig:DistingExp}
\end{figure}

\noindent\emph{Hyperparameters} For our proposed approach, we utilize the following choice of hyperparemeters: 
patch size: same as in the craft paper for vision? for language?;
number of concepts - is there a reason there is 10 per cert and uncert in experiment 1, and 55 for cert and 35 for uncert in experiment 2? Both have 10 classes?
layer - penultimate, following \cite{fel2024holistic};
backbone network - for vision tasks we use ResNet-50 and for language a BERT model

\subsection{Distinguishing Sources for Uncertainty}\label{subsec:distingUnc}

\begin{figure}[t]
\includegraphics[width=\textwidth]{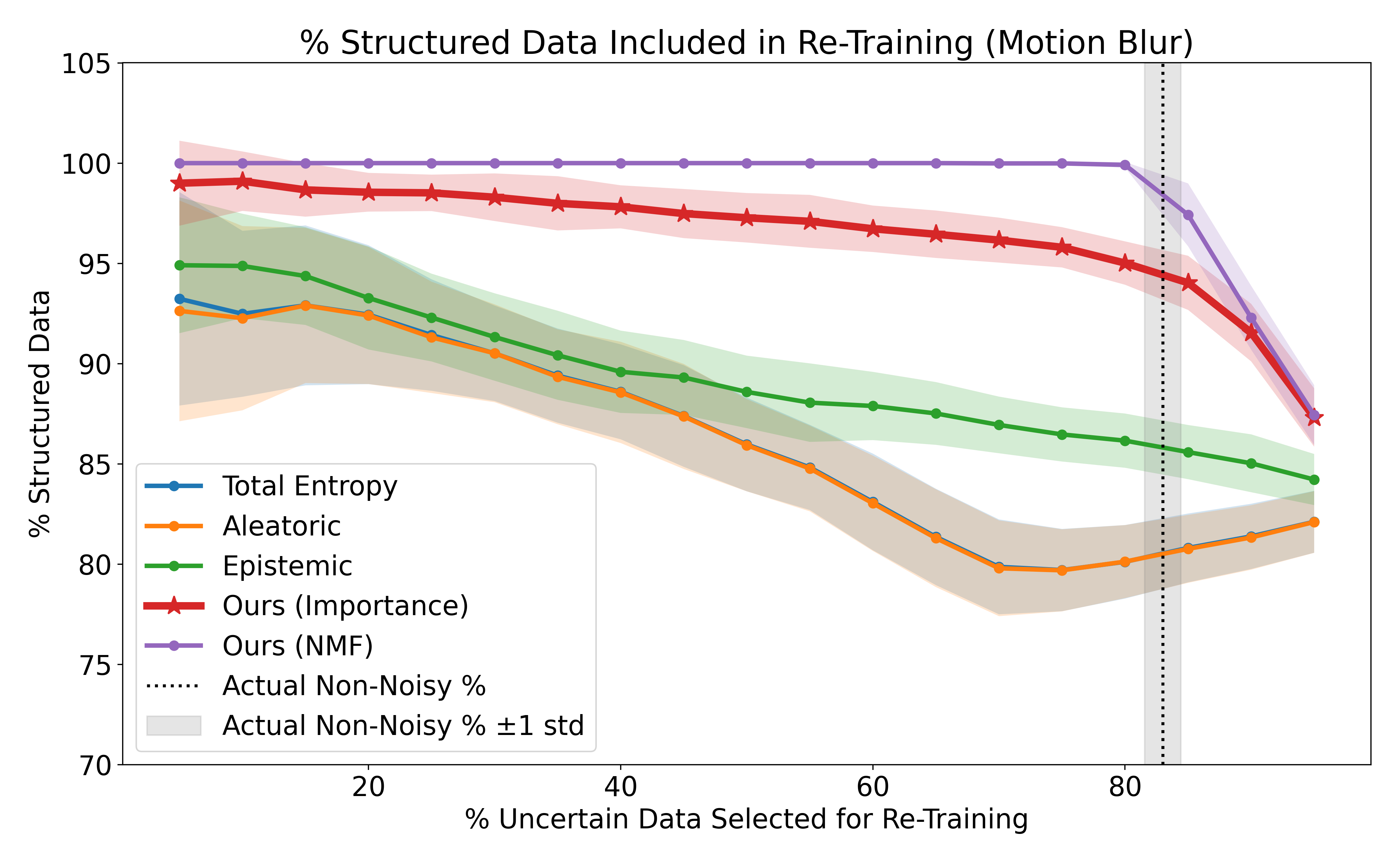}
\caption{Experiment 4.1 results for Motion Blur ($\uparrow$): When grouping uncertain data for re-training, we list the percentage of useful data among the selected ones (y-axis) for the different percentages of selected data (x-axis).} \label{fig:strnoise}
\end{figure}

In this experiment, we assess the quality and usefulness of the obtained explanations by showing that the learned concepts allow for a grouping of different sources of uncertainty, such that humans can better identify these sources and make more informed downstream decisions.

Specifically, the goal is to separate 
sources of uncertainty, and we show its benefit by building a representative re-training set in a setting where a deployed model $\model$ is exposed to two noise types: 1. Structured noise, such as the introduction of novel classes not seen during training, and 2. Unstructured noise, such as random distortions like blurring. We assume that the unstructured noise has rendered the image unusable, such that, in the context of re-training, no meaningful signal can be derived from the affected inputs.

For more clarity, we include Figure \ref{fig:DistingExp}. We begin by applying Steps 1 \& 2 from our pipeline to $\model$ to obtain the uncertain inputs. Then, with Step 3, we extract their concepts. Since concepts provide a human understanding of the source of uncertainty, a practitioner visually inspects the concepts by examining the patches most activated by each concept in order to identify those associated with the unstructured noise. In Phase 2 of Figure \ref{fig:DistingExp}, the middle concept describes a blur. Now, using our explanations, we can automatically locate the inputs that most heavily activate this concept and exclude them from our re-training set. 

More precisely, we consider the setting where new data samples $\X$ are available and our explanation pipeline is applied to provide concept banks $\V_{\cer}, \V_{\unc}$ of 10 concepts each, according local importances $\localRel(\x_i), \forall \x_i\in\X$ and global ones $\globalRelCER, \globalRelUNC$. We aim to evaluate how well $\V_{\unc}$ captures different noise types, and how well unstructured noise samples $\x_i$ can be filtered out using $\localRel(\x_i)$ and $\W_i$ of $\V_{\unc}$ concepts.
For this purpose, we assume that we can determine the set $\nc$ that corresponds to unstructured noise concepts in $\V_{\unc}$ (through visual inspection). Then, for each data point $\x_i$, we sum the local importance $\localRelNC(\x_i)$ or NMF activations $\W_{i,\nc}$ and filter out according to the highest values, corresponding to the amount of presence of these concepts. We refer to these strategies as \emph{Ours (Importance)} when using $\localRelNC(\x_i)$, and to and \emph{Ours (NMF)} for $\W_{i,\nc}$.


To implement this experiment, we use a frozen and pre-trained ResNet50 bottleneck for $g$ and train a classification head on ten dog species from the ImageWoof \cite{imagenette} dataset, a subset of ImageNet \cite{imagenet_cvpr09}. We sample 1,000 images from the test set and introduce 150 out-of-distribution (OOD) images randomly selected from the NINCO dataset \cite{bitterwolf2023outfixingimagenetoutofdistribution}. Additionally, we apply random noise to 10\% of the dog images, according to one of $\{$Gaussian noise, Salt and Pepper noise, Wave noise, Motion Blur, Gaussian Blur, and Radial Blur$\}$. We visually depict the effect of the Motion Blur in Figure \ref{fig:DistingExp} (left) and Gaussian Noise in Figure \ref{fig:example}. 
For our pipeline, we utilize $\ucurr = \utoth$ and MC Dropout and consider baselines that rank data points according to predictive uncertainty directly, filtering out first according to highest uncertainty. We utilize uncertainties based on $\utoth, \ualh, \ueph$ for the baselines.
To simulate the human-in-the-loop that visually inspects $\V_{\unc}$, we utilize a linear logistic regression classifier trained on image patches in the NMF space to identify random noise. 
We record the percentage of informative data points for different percentages of data points kept from the uncertain set. The averaged curves over 20 iterations are plotted for Motion Blurring in Figure \ref{fig:strnoise} and their respective AUCs are summarized in Table \ref{table:strnoisyauc}. 

\begin{table}[t]
\centering
\caption{We report the average AUC score $(\uparrow)$ over 20 runs for various types of noise patterns. Our proposed methods perform better than the baselines. We test our methods against Gaussian Blurring (G Blur), Salt and Pepper Noise (S and P), Gaussian Noise (G Noise), Motion Blurring (M Blur), Radial Blurring (R Blur), and Wave Noise(Wave).}
\begin{tabular}{lccccc} 
\toprule
Method & Total & Aleatoric & Epistemic & \textbf{Ours} (Imp)  & \textbf{Ours} (NMF)  \\
\midrule
G Blur & $79.6\pm 1.5$ & $79.5\pm 1.5$ & $82.6\pm 1.2$ & $\underline{85.6}\pm 1.3$ & $\textbf{89.0}\pm 0.2$ \\
SnP Noise & $77.2\pm 1.4$ & $77.3\pm 1.4$ & $76.7\pm 1.7$ & $\underline{85.4}\pm 0.9$ & $\textbf{88.9}\pm 0.2$ \\
G Noise & $82.3\pm 1.3$ & $82.5\pm 1.3$ & $70.8\pm 2.7$ & $\underline{85.1}\pm 1.2$ & $\textbf{88.8} \pm 0.3$ \\
M Blur & $77.5\pm 2.0$ & $77.4\pm 2.0$ & $80.4\pm 1.4$ & $\underline{87.0}\pm 0.9$ & $\textbf{89.2}\pm 0.2$ \\
R Blur & $82.2\pm 1.6$ & $82.4\pm 1.6$ & $73.1\pm 2.9$ & $\underline{85.2}\pm 4.4$ & $\textbf{86.2}\pm 4.9$ \\
Wave Noise  & $80.8\pm 7.4$ & $80.1\pm 8.0$ & $87.8\pm 2.0$ & $\textbf{88.3}\pm 1.5$ & $\textbf{88.3} \pm 1.5$  \\
Average & $79.9 \pm 2.2$  & $79.9\pm 2.3$ & $78.5\pm 6.3$  &$\underline{86.1}\pm 1.3$ &$\textbf{88.4}\pm 1.1$ \\
\bottomrule
\end{tabular}\label{table:strnoisyauc}
\end{table}

The results in Table \ref{table:strnoisyauc} indicate that our method outperforms uncertainty-based measures for this task. Thereby, using $\localRel(g(\x_i))$ achieves the second-best performance (underlined), while leveraging $\W_i$ yields the best results (bolded), with the highest average AUC score. Inspecting Figure \ref{fig:strnoise}, 
we can see that our method using the NMF coefficients (purple) consistently recommends images that would be beneficial to the re-training set while abstaining from recommending the unusable images until they must be chosen. We indicate the true unusable image percentage by the vertical line on the figure. In the ideal case, a method would not recommend any unusable images until it reaches the vertical line. 
Our method provides an explainable way to automatically select images that contain meaningful structure for domain adaptation.

\subsection{Rejecting Uncertain Points with Concepts}\label{exp:rej}

\begin{figure}[t]
\includegraphics[width=\textwidth]{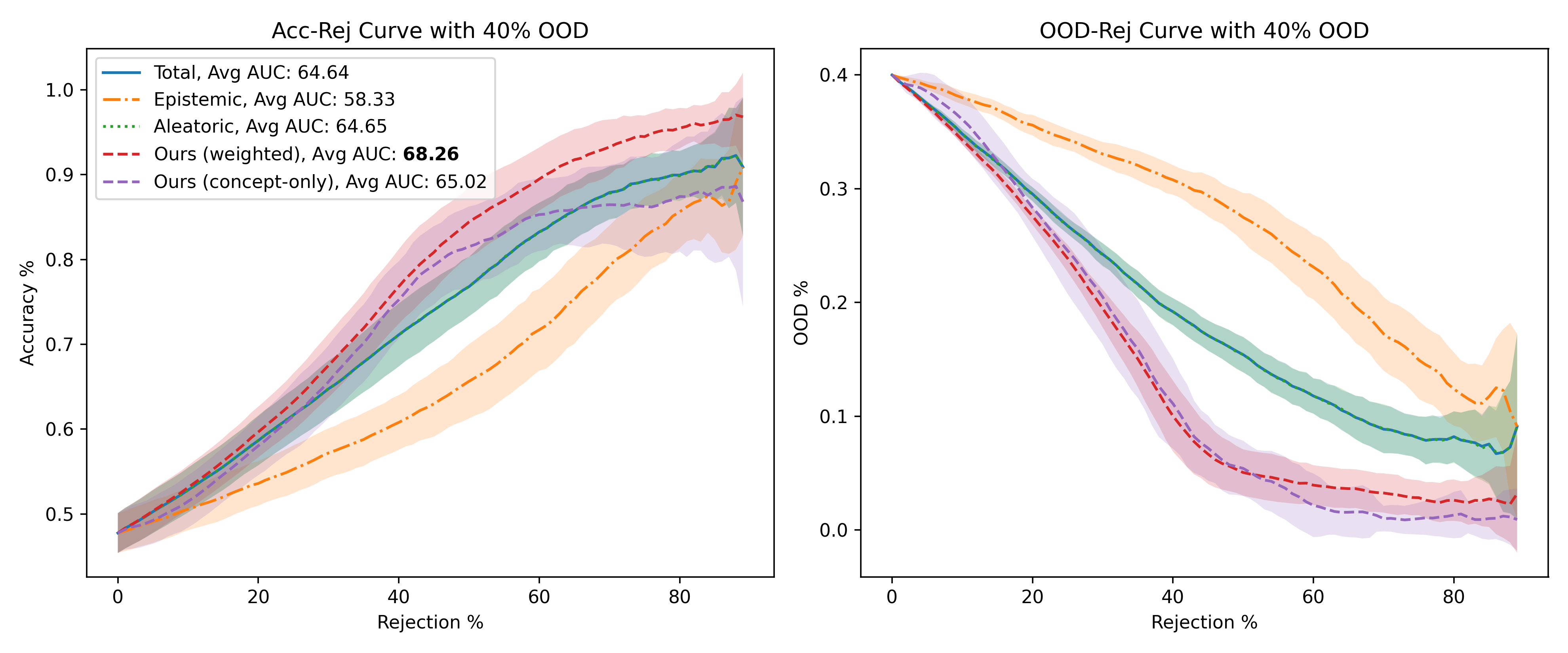}
\caption{Results for Experiment 4.2.  Accuracy-Rejection curves ($\uparrow$) show the accuracy over different percentages of rejected points (\textbf{left}). OOD-Rejection curves ($\downarrow$) show how many OOD samples remain after rejection (\textbf{right}). "Ours (weighted)" rejects more OOD points and has a higher accuracy $\geq 20\%$ rejection.} \label{fig:rejcurve}
\end{figure}

We demonstrate that our proposed explanations encapsulate uncertainty by utilizing the learned concepts in a classification setting to improve decision-making. Often,  uncertainty estimations are evaluated \textit{indirectly} by measuring the improvement of predictions \cite{H_llermeier_2021} through accuracy-rejection curves, which depict the accuracy (y-axis) of a classifier as a function of its rejection rates (x-axis) \cite{pmlr-v8-nadeem10a,10.1007/978-3-030-44584-3_35}. If the estimation performs well, we should expect the curve to be monotonically increasing. In this experiment, we create a rejection strategy using our explanations and thus evaluate their effectiveness as an uncertainty estimator. We also compare the results to baseline uncertainty estimations.
In particular, given a trained model $\model$ and a set of new data points $\X$, we apply our proposed approach to generate concept banks $\V_{\cer}, \V_{\unc}$, according local importances $\localRel(\x_i), \forall \x_i\in\X$ and global ones $\globalRelCER, \globalRelUNC$. 

Now, we build two strategies: 1. \emph{Concept-only} rejection: we identify for each input $\x_i$ the most strongly activated concept $ c^* = \arg \max \W_i$ utilizing the combined concept bank $[\V_{\cer}, \V_{\unc}]$ and determine the global importance of $c^*$. We then reject those points first, for which $c^* \in \globalRelUNC$ and with the highest $\globalRelUNC$ value.
This leads to inputs associated with globally uncertain concepts being rejected first, while those linked to globally certain concepts are retained longer. 2. \emph{Weighted} rejection: We adapt the previous strategy, by weighting the uncertainty output $f(\x_i)$ with $+1$ or $-1$, depending on whether $c^* \in \globalRelUNC$ or $c^* \in \globalRelCER$ and again reject according to highest value.

Concerning the implementation, we use a pre-trained ResNet-50 classifier as the base model. We randomly sample images from 10 out of 20 ImageNet classes (ImageWoof \cite{imagenette} + Imagenette \cite{imagenette}) and also include out-of-distribution (OOD) samples from the NINCO dataset \cite{bitterwolf2023outfixingimagenetoutofdistribution}. In each run, we use 1,000 points, with 40\% being OOD to compute the accuracy-rejection curves. We additionally compare our two concept-based strategies, Concept-only and Weighted rejects, to baselines, which reject according to highest predictive uncertainty, using each of $\utoth, \ueph, \ualh$. We also compute the AUC for each strategy and average these measures over 20 runs. We set the number of concepts to 55 for the certain group and 35 for the uncertain. 

As seen in Figure \ref{fig:rejcurve} (left), our ``weighted'' method is a monotonically increasing curve and performs with the highest AUC score computed across 20 runs. To confirm the statistical significance of the AUC score, we conduct a one-sided Wilcoxon signed rank test against the Total Entropy (which in this case is very similar to the Aleatoric curve), obtaining a p-value 
$ < 10^{-6}$, indicating ours performing significantly better. Meanwhile, the ``concept-only'' strategy is also monotonically increasing and performs better for medium rejection rates and worse for higher ones, in comparison to the baselines. 

In Figure \ref{fig:rejcurve} (right), we plot the percentage of OOD points as a function of the rejection rate.
Up to approximately 20\% rejection, all curves except the Epistemic and Concept-only one, 
exhibit similar performance. However, beyond this point, both our strategies reject a greater proportion of OOD samples at each step.
This signifies why our explanations work well as an uncertainty estimator and better than the baselines: the learned concepts pick up on the OOD data and aid in their rejection. 


\subsection{Explaining Uncertainty in Language Models for Fairness}










In this experiment, we demonstrate our proposed pipeline can also be applied to the natural language domain and that our explanations can capture sensitive group information, which can be used to correct bias in a model's predictions. 

For this purpose, we fine-tune a BERT \cite{devlin-etal-2019-bert} model on the Bias in Bios dataset \cite{10.1145/3287560.3287572} which consists of biographies where the task is to predict their corresponding occupation. Before fine-tuning, we incorporate ReLU on the last embedding layer to ensure non-negativity.  We then apply our pipeline using a NMF-based concept extraction technique for the text domain \cite{jourdan-etal-2023-cockatiel}. 
We inspect the "physician" class by 
training an NMF on the inputs predicted as such and compute their importances with respect to each group of uncertain and certain points. We investigate the most important concept of the uncertain samples, concept 6, in the following analysis. 

\begin{figure}[t]
\centering
\includegraphics[width=0.4\textwidth]{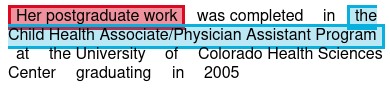}
\includegraphics[width=0.4\textwidth]{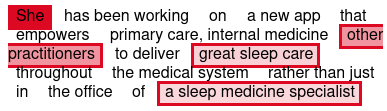}
\caption{Text excerpts where activations of concept 6 are highlighted with red.} \label{fig:equ_odds}
\end{figure}

In Figure \ref{fig:equ_odds}, to understand what concept 6 represents, we show two excerpts that activate concept 6. Thereby, the intensity of red marks the strength of the activation of concept 6.
We can see that female pronouns appear among the highlighted words along with other nouns like "co-founder" and "sleep medicine specialist". Since we know the gender labels of the dataset, we check the Pearson correlation between concept 6 and the labels. Indeed, it is the most correlated with gender at $R=0.3$. 
We further verify the relevance of this concept for representing gender by excluding it in the NMF reconstruction and applying the occupation classifier. 
This changes some of the predictions, most notably a large proportion of professors and chiropractors who were falsely predicted as physicians. Even more interesting, the gender distribution among the samples influenced by concept 6 does not align with the gender distribution of said classes, which could be an indication of gender bias in BERT.
We evaluate the change in gender bias by computing the equalized odds score before and after our intervention and report an improvement of 0.0027.
At first glance, this score does not sound impressive, but it is worth noting that the intervention on the physician class only affects a limited number of samples and thus has a limited effect on global equalized odds. More precisely, the best possible outcome for an intervention on the physician class (fixing all false positives) would have led to an improvement of 0.0069.
This demonstrates that the detected relevant concept in the uncertain group encodes gender information in our example and that removing it can improve fairness.






\section{Limitations and Future Work}\label{sec:lim}

While our approach performs well in the tasks outlined above, it is not without limitations. Concept-based explanations provide a human-interpretable means of understanding uncertainty in machine learning settings, particularly when the source of uncertainty is visibly discernible. However, they may fail to capture finer-grained pixel-level nuances of uncertainty. In this study, we maintained a fixed patch size when training the NMF, but varying the patch size could potentially capture more localized properties of uncertainty. Investigating the relationship between patch size 
and its impact on concept-based explanation quality presents a promising direction for future research. This limitation is further exemplified by changing the OOD percentage in Experiment \ref{exp:rej} from 40\% to 20\% as seen in the Appendix. We observe the convergence of the performance of our methods with the baselines. The OOD data provides clear differences captured by concepts, such that we can reject a classification based on an input's most activated concept; however, when the number of OOD is decreased, the source of the uncertainty becomes more subtle between known classes that are difficult to discriminate.

Additionally, our method for estimating concept importance may not be optimal, as evidenced by the superior performance of using NMF activations directly in our experiments. We suspect this performance gap arises from the variance, or lack thereof, of the uncertainty measure. Specifically, if we perturb a concept within an already highly uncertain input, the uncertainty measure may not exhibit significant variation. While we did not explore alternative ways to refine importance attribution in this study, we do plan to address it in future work. 

Finally, while the usefulness of our concept-based uncertainty explanations is evaluated in downstream tasks, they do
elicit downstream human action and decision-making. This prompts a user-centric study to evaluate the effectiveness of our proposed explanations in explaining uncertainty to a user. Such a study is a subject for future work.

 \section{Conclusion}
We introduced a novel framework for explaining uncertainty using automatically extracted concept activation vectors. Our proposed framework enables
both local and global explanations of uncertainty through the use of importance scores and attribution maps. These explanations demonstrate their utility by encapsulating uncertainty, aiding the design of useful re-training sets, incorporating them into rejection strategies, and helping to detect and mitigate bias. Moreover, while concept-based explanations of model predictions can be useful, using CAVs to capture sources of uncertainty not only offers another complementary view into how a model makes its decisions but also provides interpretable ways to enhance its performance.

\section{Acknowledgements}
This project has received funding from the European Union's Horizon Europe research and innovation programme under the Marie Sk\l{}odowska-Curie grant agreement No 101073307.

%
%
%
\bibliographystyle{splncs04}
\bibliography{bibliography}
%




\section{Appendix}

\begin{figure}[H]
\includegraphics[width=\textwidth]{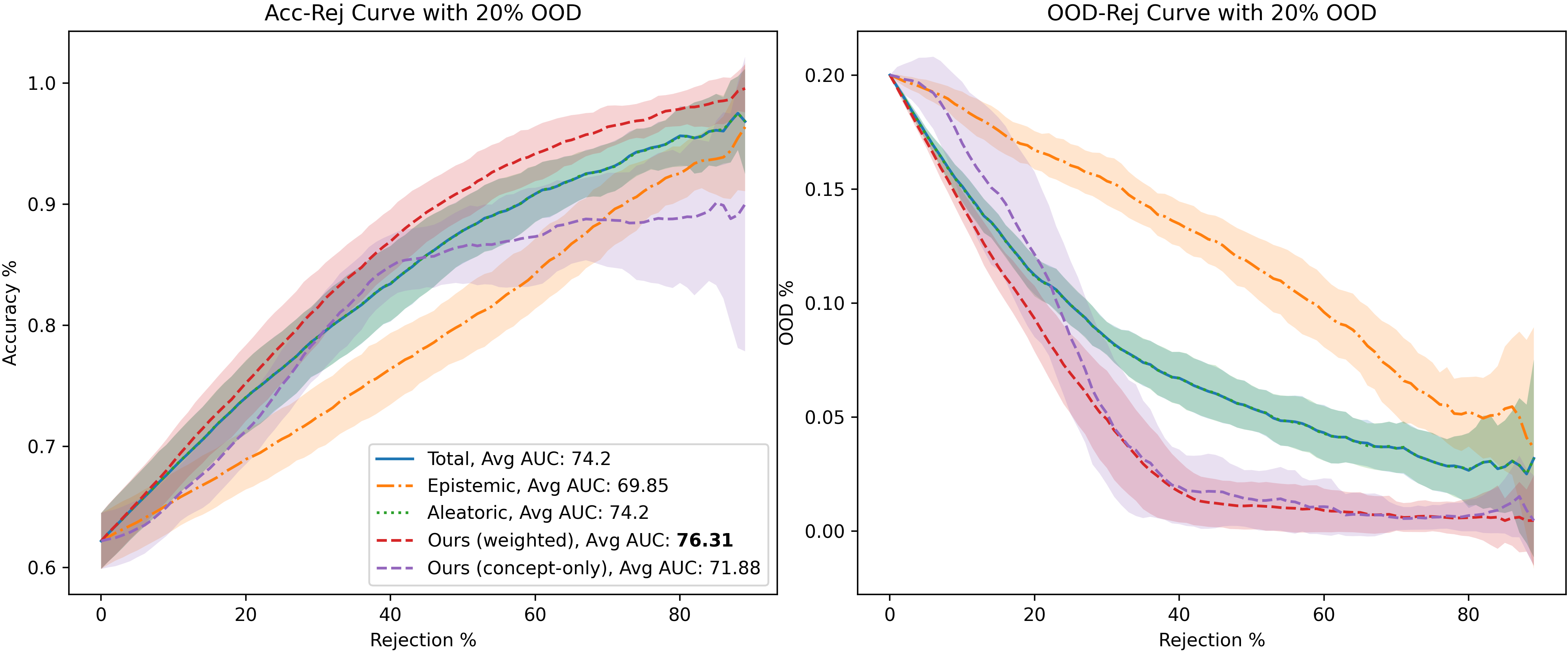}
\caption{(\textbf{left}) Accuracy-Rejection ($\uparrow$) and (\textbf{right}) OOD-Rejection Curves ($\downarrow$) with 20\% OOD data. } \label{fig:rejcurve20}
\end{figure}

For Figure \ref{fig:rejcurve20}, we use the setup described in Experiment \ref{exp:rej} except we include only 20\% OOD data. Our ``weighted'' method is a monotonically increasing curve and performs with the highest AUC score computed across 20 runs, which is confirmed by a one-sided Wilcoxon signed rank test against the Total Entropy, obtaining a p-value 
$ < 9^{-7}$. Meanwhile, the ``concept-only'' strategy monotonically increases until only around a 40\% rejection rate. On the OOD-Rejection curve (right) at 40\% rejection, both strategies have rejected nearly all of the OOD data, leaving only in-distribution data. However, we can see that the combination of the uncertainty value given by our pipeline and the use of concepts (our ``weighted'' method) produces a better curve, suggesting a synergy between uncertainty quantification and concepts.

\begin{figure}[t]
\includegraphics[width=\textwidth]{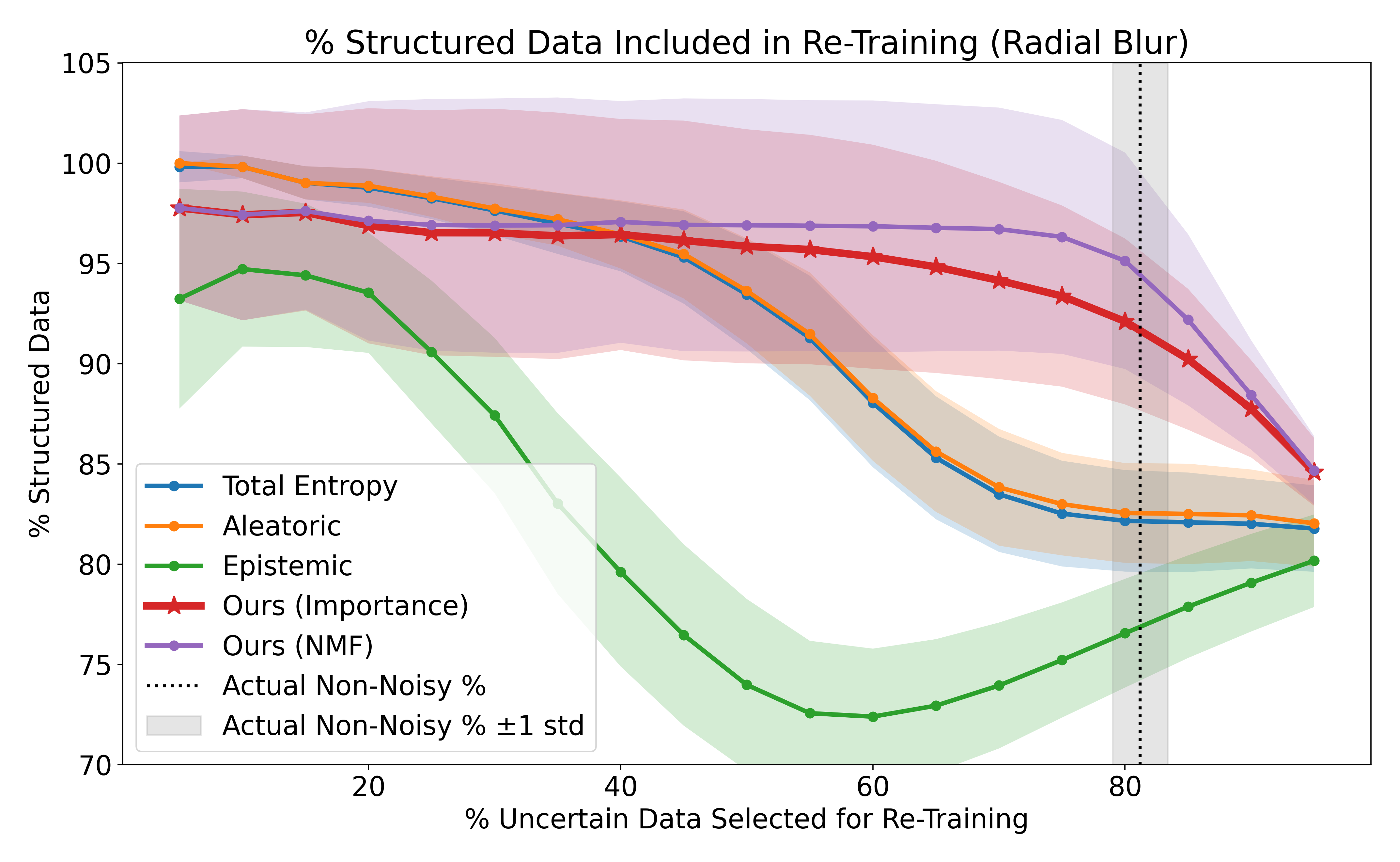}
\caption{(\textbf{left}) Radial Blur experiment results ($\uparrow$).} \label{fig:strnoiseradial}
\end{figure}

\begin{figure}[t]
\includegraphics[width=\textwidth]{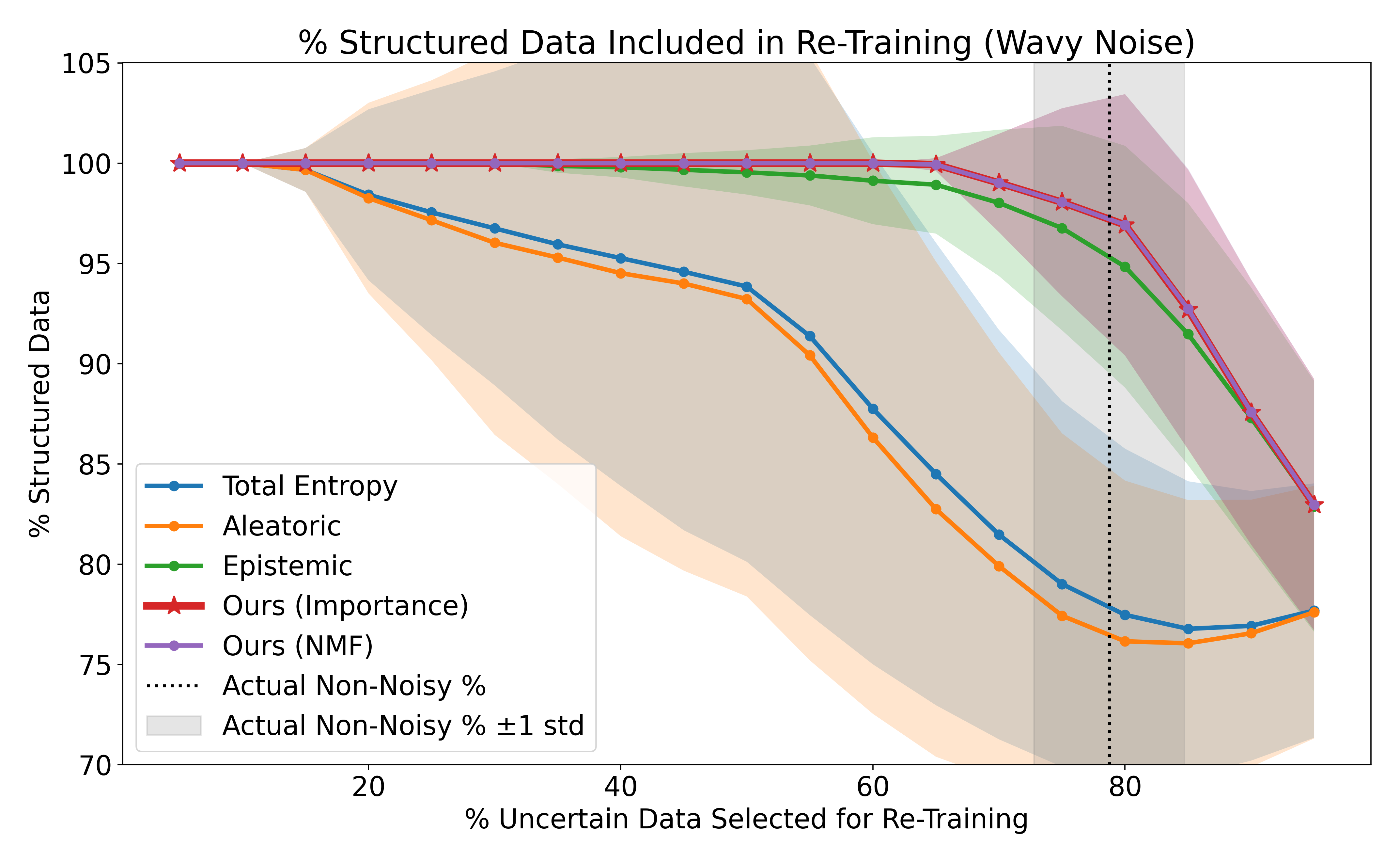}
\caption{(\textbf{left}) Wave Noise experiment results ($\uparrow$).} \label{fig:strnoisewave}
\end{figure}

\begin{figure}[t]
\includegraphics[width=\textwidth]{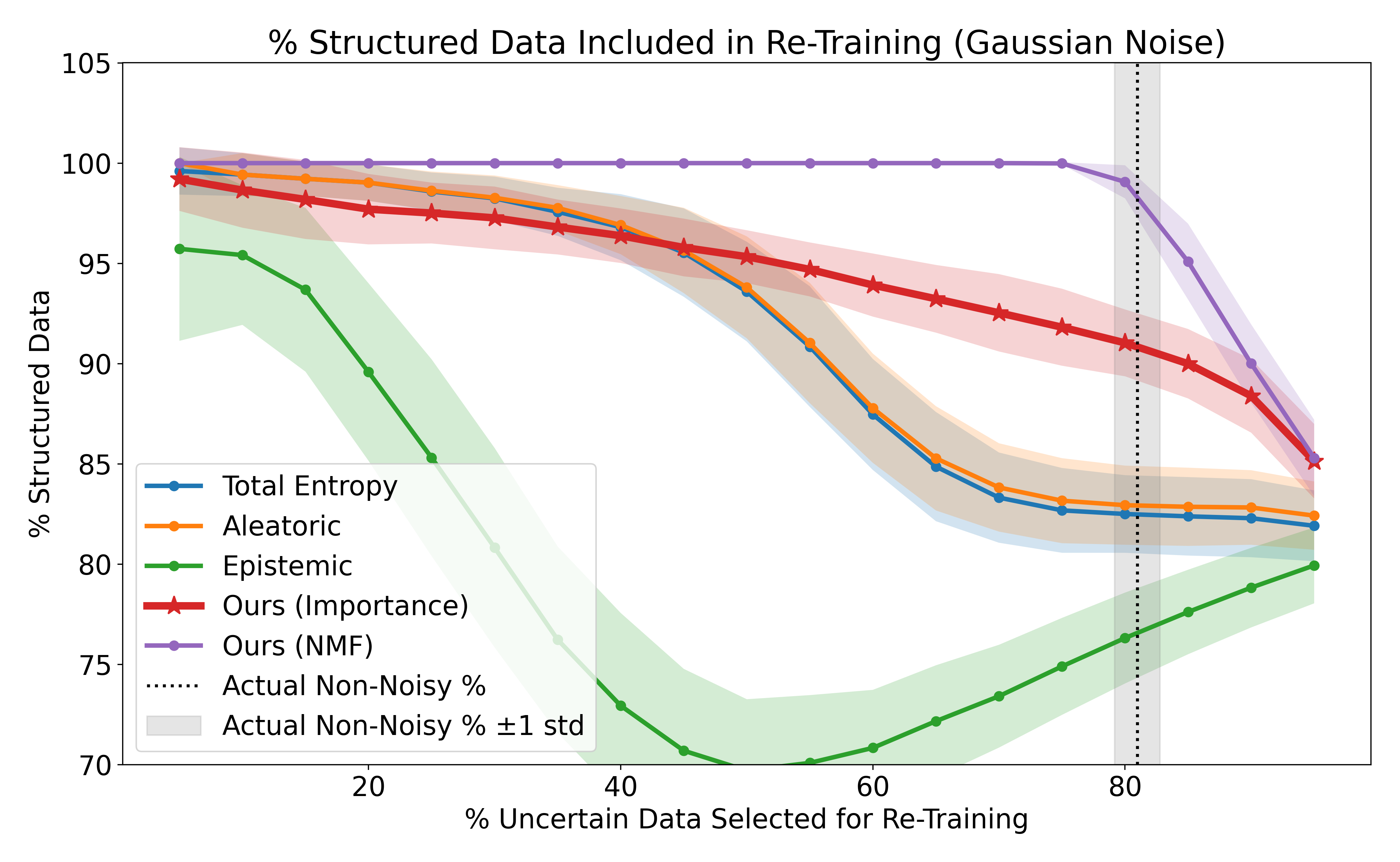}
\caption{(\textbf{left}) Radial Blur experiment results ($\uparrow$).} \label{fig:strnoisegn}
\end{figure}


\begin{figure}[t]
\includegraphics[width=\textwidth]{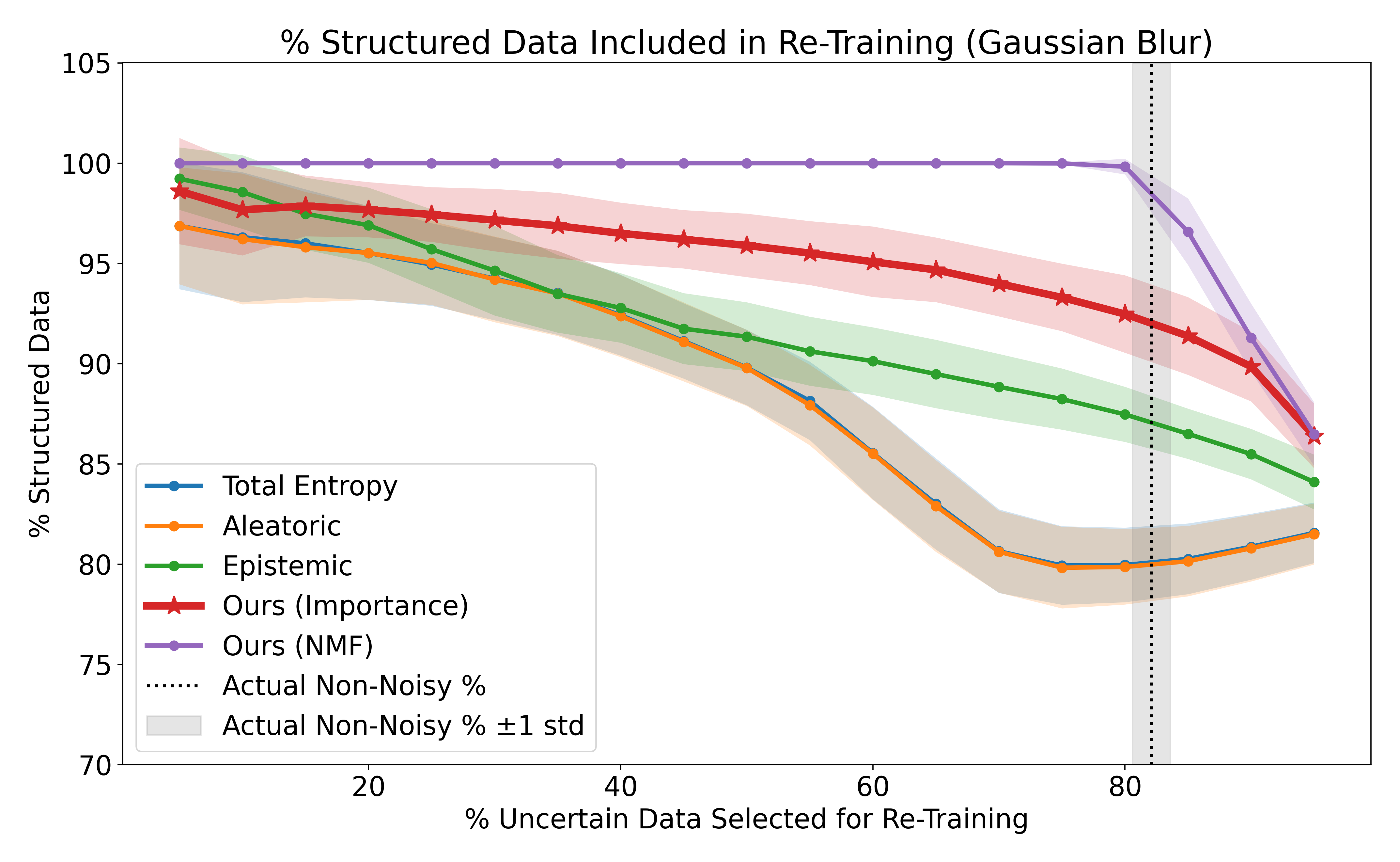}
\caption{(\textbf{left}) Gaussian Blur experiment results ($\uparrow$).} \label{fig:strnoisegblur}
\end{figure}

\begin{figure}[t]
\includegraphics[width=\textwidth]{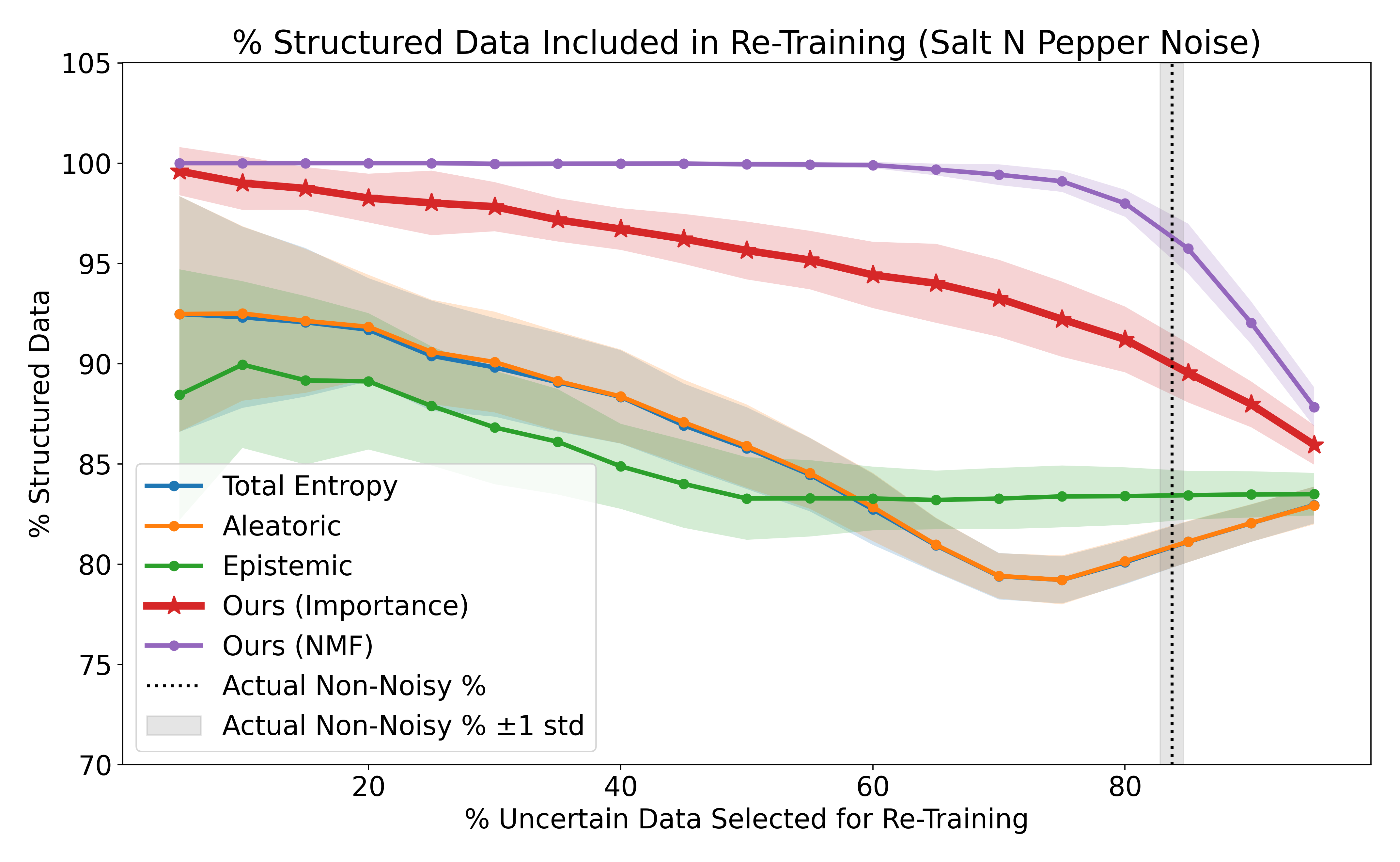}
\caption{(\textbf{left})  Salt and Pepper noise experiment results ($\uparrow$).} \label{fig:strnoisesaltnpepper}
\end{figure}

\end{document}